\documentclass[letterpaper, 10 pt, conference]{ieeeconf}  

\usepackage[utf8]{inputenc}
\usepackage[T1]{fontenc}
\usepackage{verbatim}
\usepackage{graphicx}
\usepackage{caption}
\usepackage{subcaption}
\usepackage{wrapfig}
\usepackage{amsmath}
\usepackage{biblatex}
\usepackage[dvipsnames]{xcolor}

\newcommand{\customfigBracket}[1]{\textit{(Figure \ref{#1})}}

\IEEEoverridecommandlockouts                              

\begin{document}
\title{\LARGE \bf
	A Soft Robotic Gripper with Active Palm for In-Hand Object Reorientation
}

\author{Thomas Mack, Ketao Zhang, Kaspar Althoefer, \textit{Senior Member, IEEE}
    \thanks{This work was supported in part by an Alan Turing Institute funded project on Intuitive human-robot interaction in work environments.}
    \thanks{The first author is funded by an iCASE EPSRC PhD studentship.}
    \thanks{For the purpose of open access, the authors have applied a Creative Commons Attribution (CC BY) license to any Accepted Manuscript version arising.}
    \thanks{Authors are with the Centre for Advanced Robotics @ Queen Mary, School of Engineering and Materials Science, Queen Mary University of London, United Kingdom.
        }%
    }
\maketitle
\thispagestyle{empty}
\pagestyle{empty}

\begin{abstract}

The human hand has an inherent ability to manipulate and re-orientate objects without external assistance. As a consequence, we are able to operate tools and perform an array of actions using just one hand, without having to continuously re-grasp objects. Emulating this functionality in robotic end-effectors remains a key area of study with efforts being made to create advanced control systems that could be used to operate complex manipulators. In this paper, a three fingered soft gripper with an active rotary palm is presented as a simpler, alternative method of performing in-hand rotations. The gripper, complete with its pneumatic suction cup to prevent object slippage,  was tested and found to be able to effectively grasp and rotate a variety of objects both quickly and precisely.



\end{abstract}

\section{Introduction}

	The act of repositioning objects within our grasp comes so naturally to humans that we often barely notice ourselves doing it. When lifting a pen we usually have to reorient it before we start writing - a precise action that most humans are able to execute in half a second. Some are even able to deftly twirl pens and similar objects between their fingers. Removing the pen lid is more challenging, but most people are able to accomplish this task single-handedly using three fingers to grip the pen and the thumb and index finger to lift the lid off. While writing, the fingers accurately move the nib in a plane to trace out letters legible enough to be read by others.

 \begin{figure}
    \centering
    \includegraphics[width=0.45\textwidth]{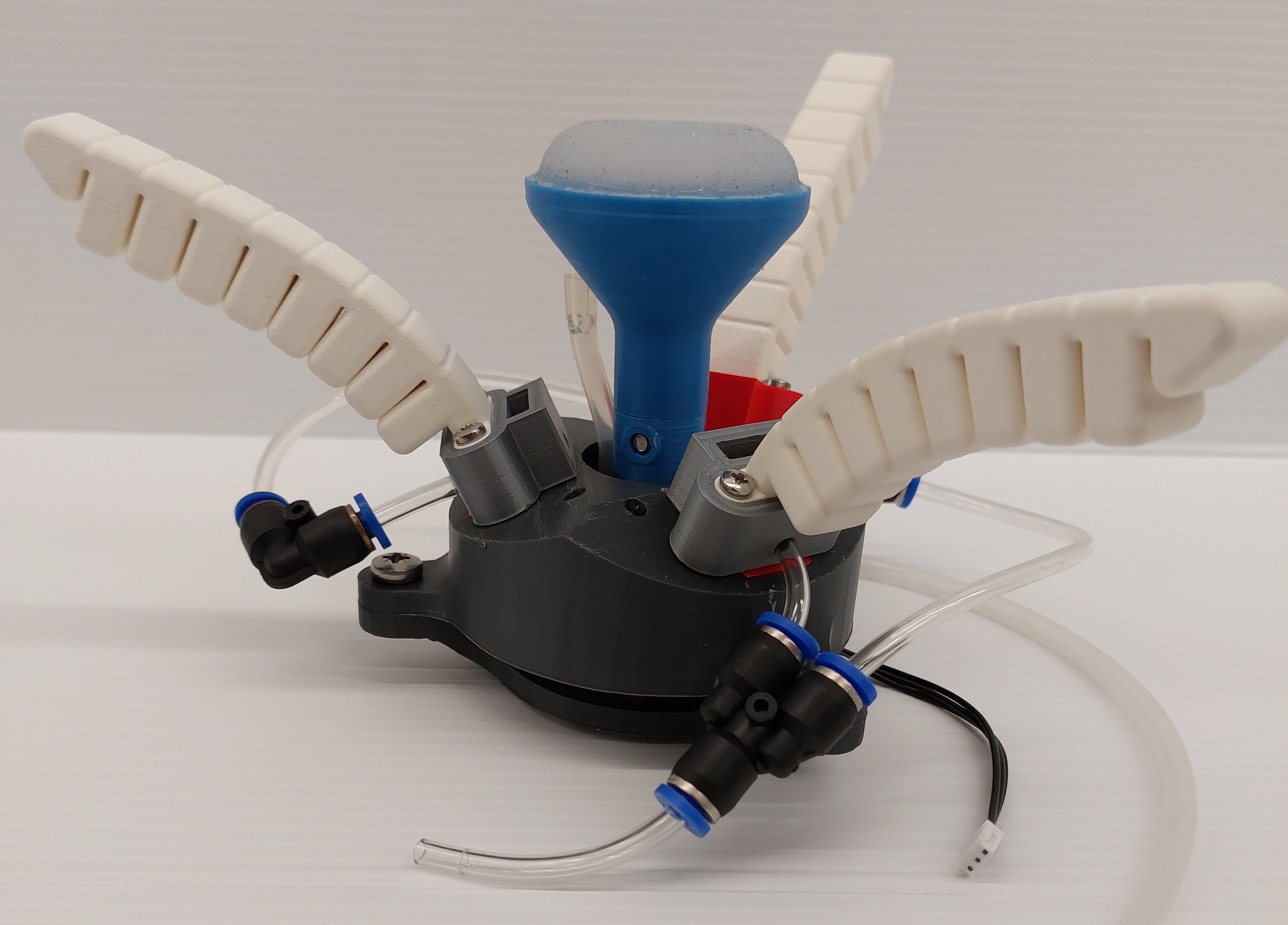}
    \caption{The gripper using the soft, 3D printed fingers
    }
    \label{fig:title}
\end{figure}

	Replicating this level of dexterity, either fully or partially, in a robotic manipulator is clearly a huge challenge for the industry, on account of the myriad of applications it would proffer. Assembly lines and packing robots would benefit enormously from this technology as items with arbitrary shapes and weights could be easily reoriented by the gripper. Potentially it could even remove the need for specialised manipulators for different stages of production.
	
	Most environments and implements encountered in everyday life are designed for humans use. Underlying any design is an inherent assumption that the humans working in those environments or using those implements will have the level of dexterity that comes so naturally. A general purpose collaborative robot would, of course, benefit from human-like dexterity in most basic tasks such as opening doors, or reorienting objects to make them easier for a human to receive. Advanced generalised grippers would be able to handle tools, enabling them perform the same tasks as humans without the need of specialised end effectors.
	
        %
	    
    
    During interactions with a robot, a human would find it most useful if objects, such as tools, were presented to them in a convenient manner. A tool, when picked up by a robot gripper with the intention to pass this on to a human, would rarely be in an orientation that can be received easily and safely.
    
    Here, a robotic system capable of rearranging the grasped tool (e.g., a screwdriver) such that when the handover to the human occurs, the human can receive it handle first, to conduct the task (e.g., fastening screws). It would immediately, make a difference in many industrial settings such as manufacturing, assembly, maintenance and inspection.
    
	The development of robotic in-hand manipulation would also enhance prosthetics and other assistive technologies. Current commercial robotic prosthetic hands are very expensive and rely mostly on an external interface setting different poses that can be proportionally controlled or toggled \cite{prosthetics} using electromyography (EMG) \cite{emg}. However, the muscles required for EMG are not always intact. Even if finger control via EMG is possible, it may not offer the precision required to manipulate objects within the hand. A method of automatically grasping everyday objects would be a great enhancement in terms of quality of life, eliminating the need to changing the grasp pose for each different object. Indeed the ability to reorient an object within the grasp using a simple control scheme would be the next step and an ideal solution if only a limited number of EMG inputs were available.
	
	Many different approaches to creating grippers capable of in-hand manipulation have been adopted in the past. As the most prominent example of a high precision, fast manipulator is the human hand, many anthropomorphic designs have been created that try to closely mimic it \cite{shadowhand}\cite{RBO3}\cite{floppy} or that are at least inspired by it. However, their inherent complexity makes them difficult to build. A human hand has over 23 degrees of freedom \cite{anatomy} and their positions and torque can be sensed with inbuilt proprioception \cite{proprioception}. Tendons are often used for actuation in robotic hands as fitting so many actuators and force and position sensors in to the fingers is almost impossible when working at a scale similar to a human hand at least with today's technology; using tendons, sensors can be situated remotely, e.g., in the arm that the gripper is mounted on.
	
	Human skin is a soft, thin, waterproof covering filled with biological sensors that provide us a wealth of information when grasping. Robotic skin is a field of study in its own right \cite{skin} and has many applications beyond in-hand manipulation. Our skin has directional force sensors that let us gauge not just the weight of an object but the torque that it exerts on our hands, which we can use to infer its centre of mass - a function which has been implemented in some grippers \cite{torqueBased}. These force sensors can also detect texture when moved over a material. Combined with the ability to sense temperature, a human can make a reasonable estimate as to the material properties through touch alone. This would all be useful information to a robotic gripper, potentially allowing it to adjust its grasp for individual objects without visual information.
	
	Controlling such a complex system to perform in-hand manipulation presents its own set of challenges. While the human hand can be used as an example for manipulation strategies, it is computationally expensive to model an object with five distinct points of contact. Most finger based in-hand manipulation is performed with the fingertips grasping objects from the side, incurring a risk of slipping which would need to be mitigated by the robot. To do this it would need to have knowledge of the object's shape and properties to successfully plan movements. This information may need to be gathered using computer vision or through contact sensors on the robot which adds another layer of complexity.

	Non-anthropomorphic systems have also been explored as simpler solutions to in-hand manipulation \cite{mccann}\cite{twoFinger}\cite{paralell}. They will often use fewer fingers or non-standard methods to manipulate objects. Instead of trying to achieve a wide variety of manipulation tasks, these grippers may be designed to effectively perform just one or a few tasks. Using fewer fingers makes the robot much easier and cheaper to build and reduces the computational load when modelling and planning movements. However, using the minimum number of fingers for manipulation means there are no extra redundant fingers that can be used to adapt the grasping style mid reorientation. As few as two fingers can achieve limited human-like in-hand rotation and translation \cite{twoFinger}.
	
	The recent advance in soft robotics has prompted its incorporation into in-hand manipulation. A soft fingertip will deform during a grasp, distributing the force over a wider area which can reduce the risk of damaging fragile objects. Materials like silicone which are often used in soft robotics also increase friction, reducing the risk of slipping. Whole fingers can be made of soft materials with chambers that can be pressurised to cause continuous bending throughout the structure \cite{abondance}\cite{continuous}\cite{pagoli}, similar to a continuum robot \cite{kasparGift_2017}. These compliant fingers can be designed to grasp a wide variety of shapes as they deform and adapt if too much air pressure is applied. This means soft grippers can often grasp objects without needing their exact dimensions - a key advantage. As the applied force is distributed, fragile objects will also tolerate more excessive force than they would with a rigid gripper. However, deformable materials and the use of pneumatics can introduce non-linearities and make soft actuators more difficult to control, sometimes requiring the use of machine learning \cite{softML}. They can also be prone to leaks and can be easily damaged.
	
	Rigid solutions \cite{mccann}\cite{paralell} are still being explored as the actuators and construction methods are well established. The kinematics are comparatively simple and can allow for far greater precision than an entirely soft robot. However, the rigidity requires precise knowledge of the manipulated object and the system must control the trajectories of all the fingers to ensure stable movement without excessive force.


\section{Conceptual Design of the 3-Fingered Soft Gripper}
    Some of the more successful grippers utilise a palm built into them for stability. \textcite{abondance} employed a palm which acted as a flat surface on which objects were rotated and translated. \textcite{pagoli} used an in-built suction cup to support objects while the fingers release and re-position them. In both cases, as with many others, these palms are in an effective place to enable in-hand rotation as they lie in the axis of rotation. They can also have adhesive materials or devices attached to stabilise the objects they handle.
    
    Here, a gripper is proposed that utilises a rotating palm with an integrated suction cup for in-hand rotation. As with other solutions, the palm is located centrally, and is surrounded by fingers used for grasping.
    
\begin{figure}
        \centering
        \begin{subfigure}[b]{0.15\textwidth}
            \includegraphics[width=\textwidth]{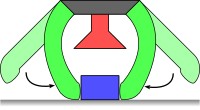}
            \caption{}
            \label{fig:process:1}
        \end{subfigure}
        \hfill
        \begin{subfigure}[b]{0.15\textwidth}
            \includegraphics[width=\textwidth]{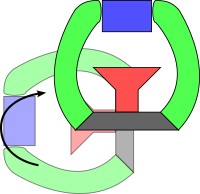}
            \caption{}
            \label{fig:process:2}
        \end{subfigure}
        \hfill
        \begin{subfigure}[b]{0.16\textwidth}
            \includegraphics[width=\textwidth]{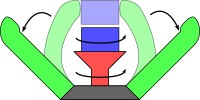}
            \caption{}
            \label{fig:process:3}
        \end{subfigure}
        \caption{The process by which objects are rotated. The chassis is shown in grey, the fingers in green and the palm in red: a) The fingers grasp and lift the object; b) The gripper and object are turned 180$^\circ$ to face up; c) The object is dropped on to the palm for rotation.
        }
        \label{fig:process}
    \end{figure}

    The normal use case for this gripper would not involve the fingers during rotation. They would be used to initially grasp the object from above. The entire gripper would then rotate 180$^\circ$ to place the palm below the object. The object would then be released and dropped on to the palm so that it could then be rotated to the desired position. The suction cup would help secure the object to the palm during rotation. The object would then be re-grasped for placing back down onto a surface \customfigBracket{fig:process}.
    
    Using a motor for rotation would allow an object to be turned quickly and precisely. It would also be largely unaffected by interactions between fingers and the object's geometry during rotation in the way that finger based manipulators are. Another advantage over finger based rotation is that no re-grasping needs to take place to achieve large changes in angle, allowing the target orientation to be reached in a single smooth twist.
    
    Because the fingers are not the source of rotation they only need to be able to reliably grasp objects. Soft pneumatic fingers are ideal as they can achieve effective grasping using a minimal control scheme \cite{continuous}. They could simply be designed to open and close, utilising their compliance rather than any specific knowledge of the objects' dimensions. This would however limit the freedom of movement of the fingers, preventing the gripper from translating objects.
    
    The proposed in-hand rotation could still be used in many applications such as in human-robot interaction. Objects could be re-positioned in the manipulator to render them easier to receive by a human. The fingers would also be soft, making any contact with a human safer. This kind of in-hand rotation could also be very useful for delicate tasks like fruit picking or packing. The only force being applied to the fruit would be from suction cup, and that would only be a marginal force, just enough to prevent the fruit from falling off the palm.

\section{Design and Prototyping}
    
    \begin{figure}
        \centering
        \begin{subfigure}[b]{0.2\textwidth}
            \centering
            \includegraphics[width=\textwidth]{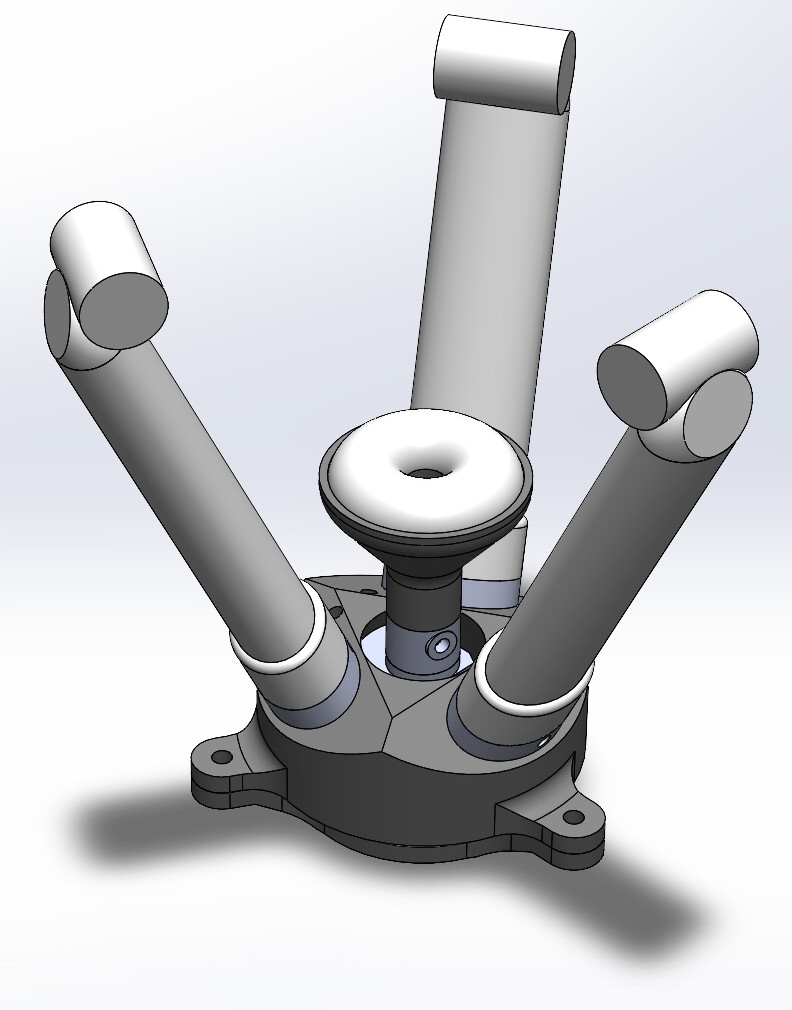}
            \label{fig:full:full}
            \caption{}
        \end{subfigure}
        \hfill
        \begin{subfigure}[b]{0.26\textwidth}
            \centering
            \includegraphics[width=\textwidth]{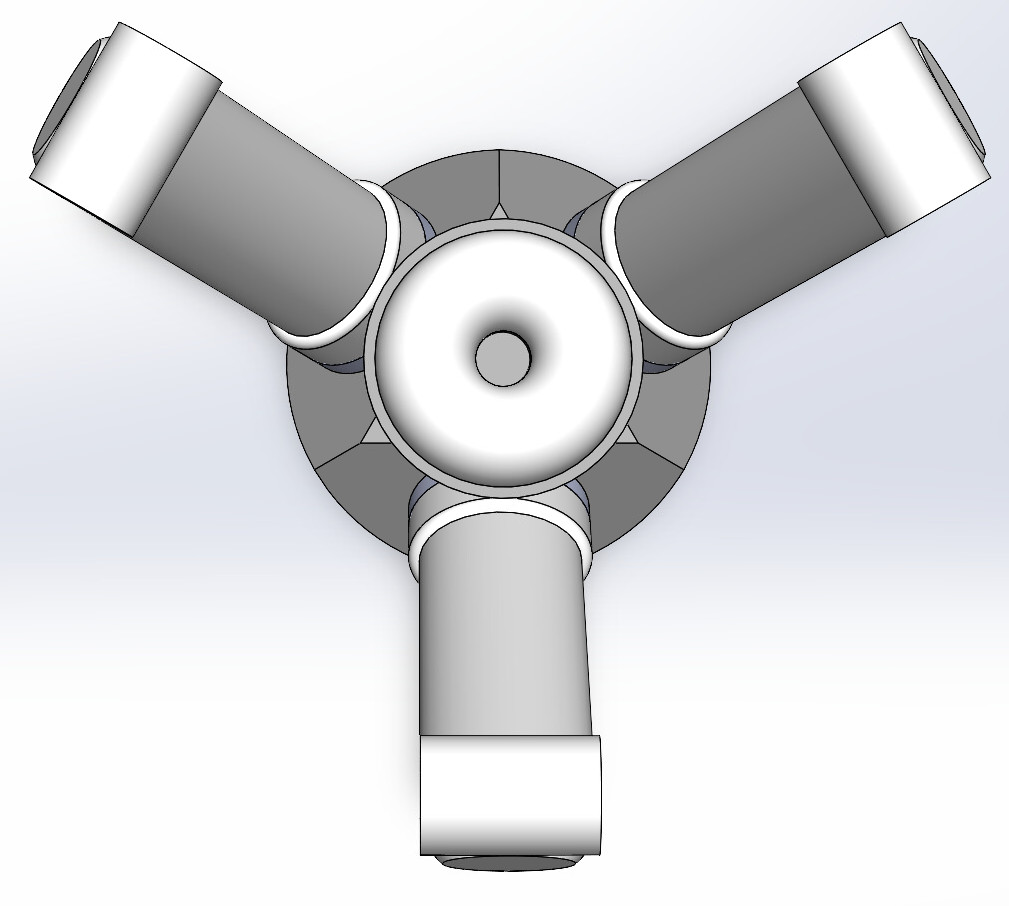}
            \label{fig:full:top}
            \caption{}
        \end{subfigure}
        \label{fig:full}
        \caption{The CAD model shown of the gripper: a) Isometric view; b) Top view
            }
    \end{figure}
    
\begin{figure}
    \centering
        \begin{subfigure}[b]{0.23\textwidth}
            \centering
            \includegraphics[width=\textwidth]{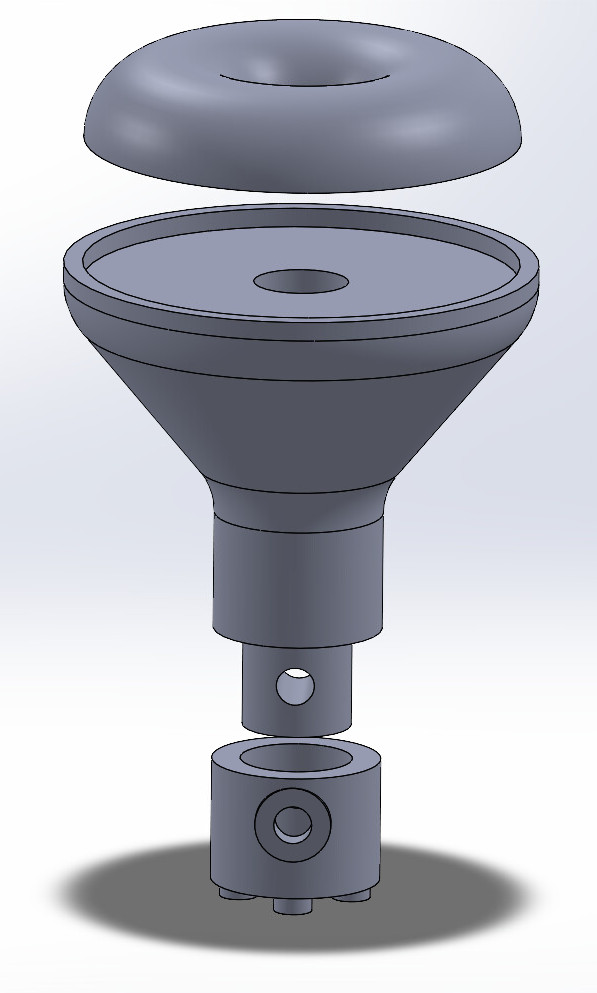}
            \caption{}
            \label{fig:palm:full}
        \end{subfigure}
        \hfill
        \begin{subfigure}[b]{0.23\textwidth}
            \centering
            \includegraphics[width=\textwidth]{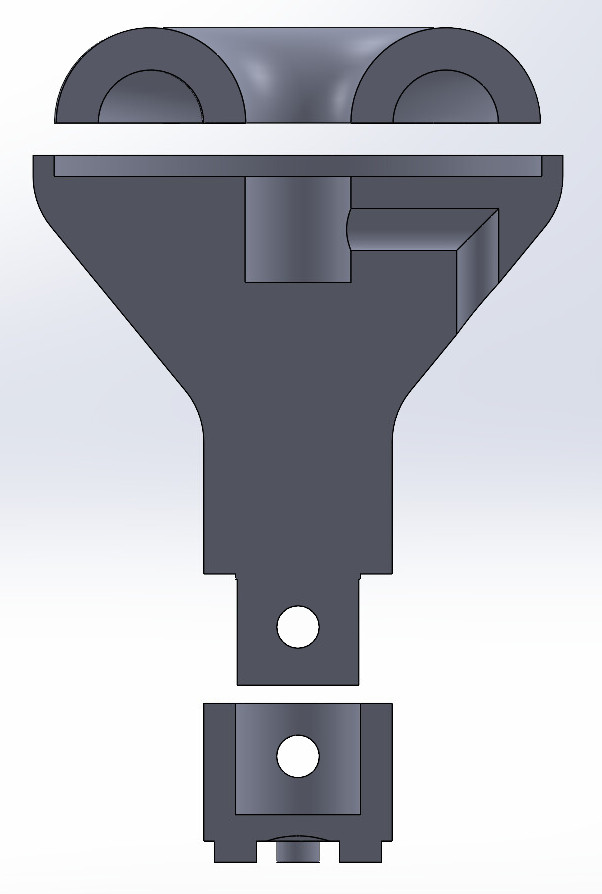}
            \caption{}
            \label{fig:palm:cut}
        \end{subfigure}
        \caption{From the top can be seen the soft silicon suction cup, the palm with integrated tubing and a piece that attaches it to the servo motor: a) Exploded; b) Exploded cross section.
        }
        \label{fig:palm}
    \end{figure}

    The gripper uses a motorised platform for the palm with an integrated pneumatic suction cup \customfigBracket{fig:palm}, attached directly at the base to an XL-320 servo motor by ROBOTIS. The suction cup is moulded from the softest variety of Ecoflex silicone from SmoothOn in the shape of a half torus shell. It is hollow to make it more deformable so as to form a better seal with non flat objects. A channel that passes from the inside of the suction cup to below the outside of the palm is designed house a silicone tube that flexes as the palm twists \customfigBracket{fig:palm:cut}. This tube is fed through the chassis and used to propagate the vacuum that is toggled with a valve. 
    
    The palm is then surrounded by three soft pneumatic fingers for grasping that all flex simultaneously. They are mounted on the chassis and spread out at 25$^\circ$ to increase the size of objects that they are able to handle.
    
    Different types of continuous pneumatic silicone fingers were designed and evaluated to find an optimal version that would allow most objects to be grasped. The soft nature of the fingers meant that they would naturally adapt to different sizes and shapes, which then left maximum liftable weight as the principal challenge.
    
\begin{figure}
        \begin{subfigure}{0.35\textwidth}
            \centering
            \begin{subfigure}[b]{\textwidth}
                \centering
                \includegraphics[width=\textwidth]{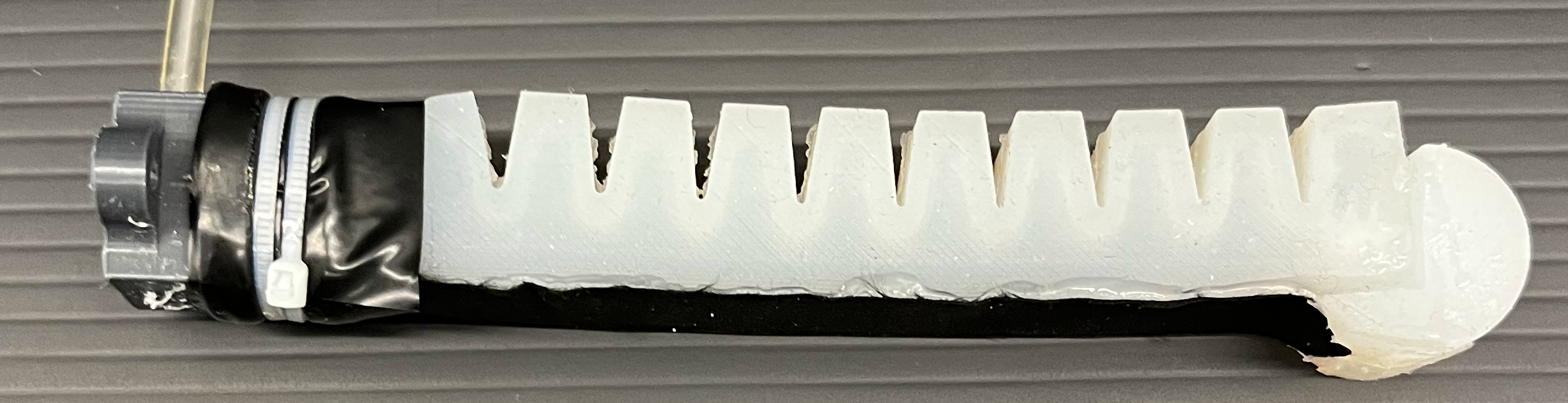}
                \caption{}
                \label{fig:finger:fabric}
            \end{subfigure}
            \hfill
            \begin{subfigure}[b]{\textwidth}
                \centering
                \includegraphics[width=\textwidth]{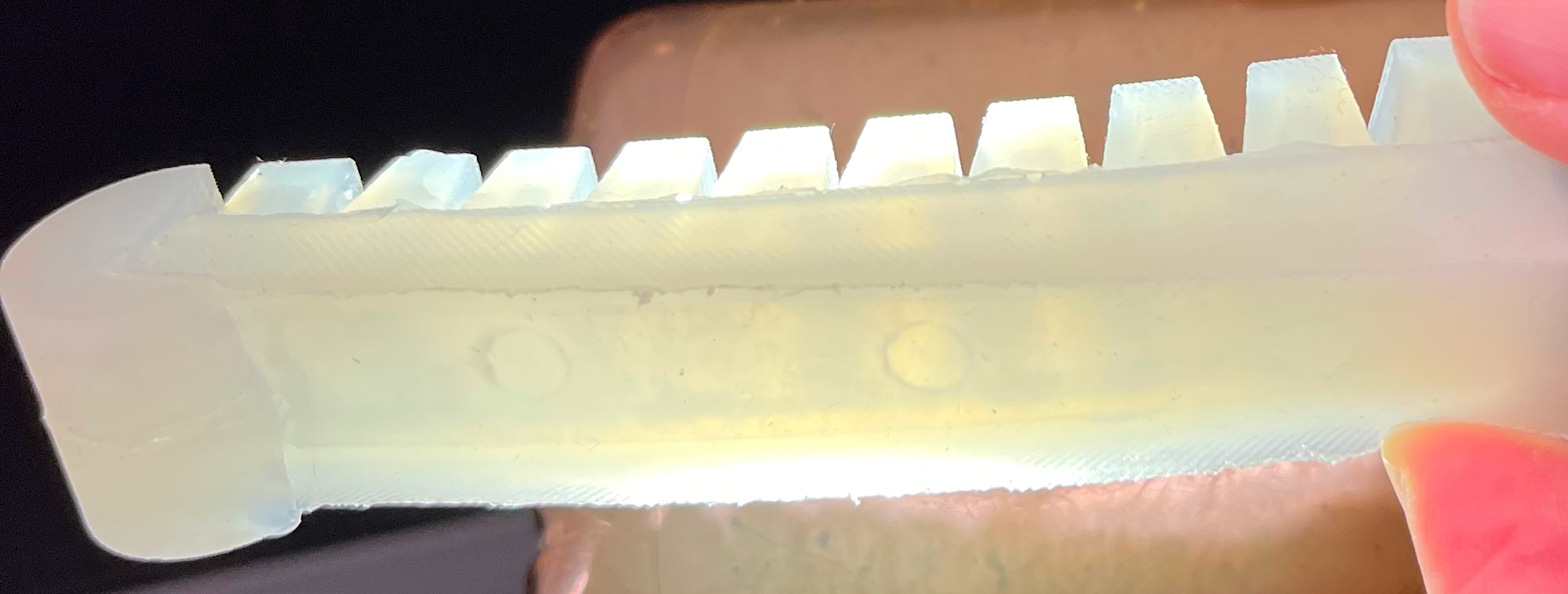}
                \caption{}
                \label{fig:finger:plastic}
            \end{subfigure}
        \end{subfigure}
        \hfill
        \begin{subfigure}[b]{0.11\textwidth}
            \centering
            \includegraphics[width=\textwidth]{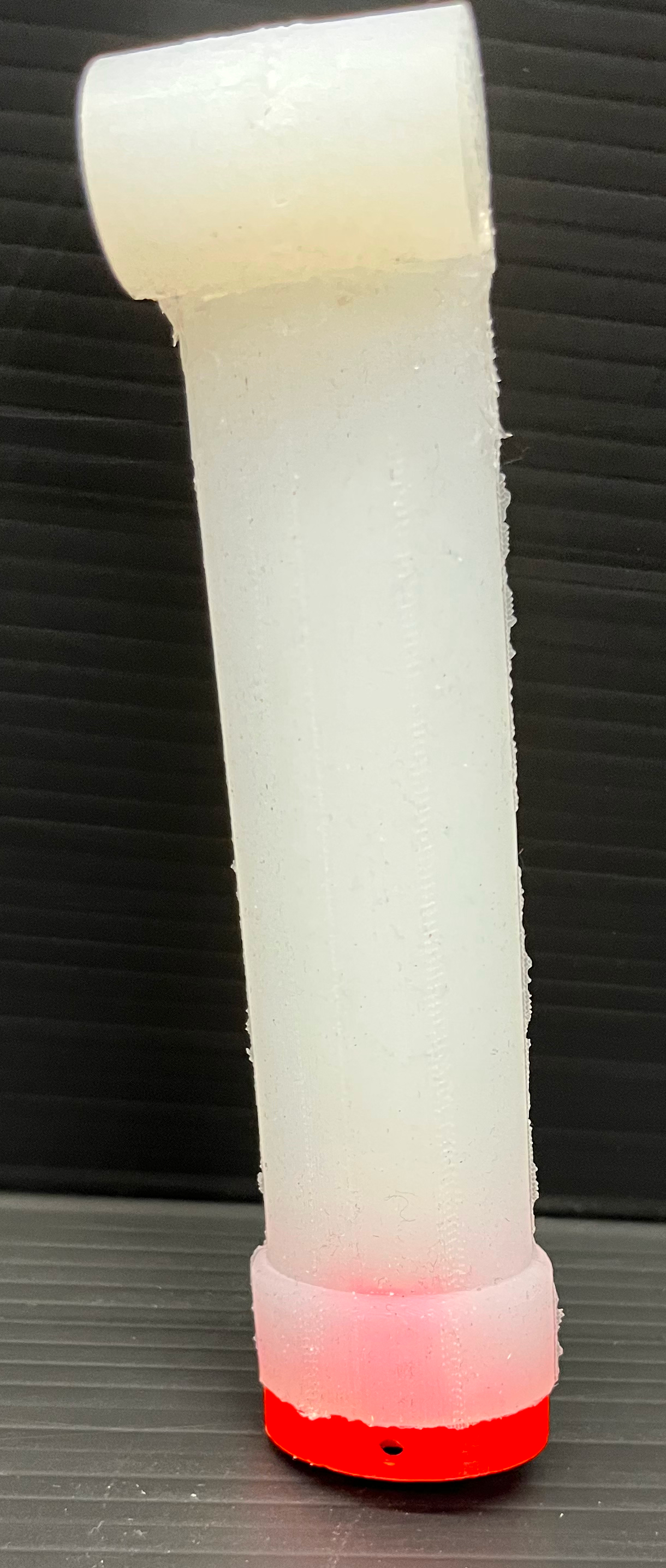}
            \caption{
            }
            \label{fig:finger:oval}
        \end{subfigure}
        \caption{Each finger designed and moulded for the gripper: a) The fabric reinforced finger; b) The plastic reinforcement can be seen in the second finger when illuminated; c) The oval shaped finger.
        }
        \label{fig:finger}
    \end{figure}

    Each finger was moulded from Dragonskin 30. A pneu-net type design inspired by \textcite{abondance} and softroboticstoolkit.com \textcite{toolkit} was first implemented, with a single set of chambers that inflate to continuously flex the finger \customfigBracket{fig:finger:fabric}. The actuators were fabricated in a single step using a 3D printed, Polylactic Acid (PLA) mould and a Polyvynil Acetate sacrificial insert. Because pneu-net style actuators often flex to extreme angles and curl over 180$^\circ$, the fingertips were made cylindrical. This was to encourage the fingertips to roll over objects as the fingers curled, bringing it closer to the palm. A layer of fabric was glued to the inside edge of the finger to prevent extension. However, when grasping was attempted, they would often deform laterally and rotate past objects, dropping them.
    
    An attempt was then made to stiffen the finger laterally. The design was modified to have a thicker base in which a polyethylene strip could be embedded \customfigBracket{fig:finger:plastic}. Both of these increased the stiffness of the finger, and the polyethylene strip resisted lateral deformation.

    \begin{figure}
        \centering
        \begin{subfigure}[b]{0.47\textwidth}
            \centering
            \includegraphics[width=\textwidth]{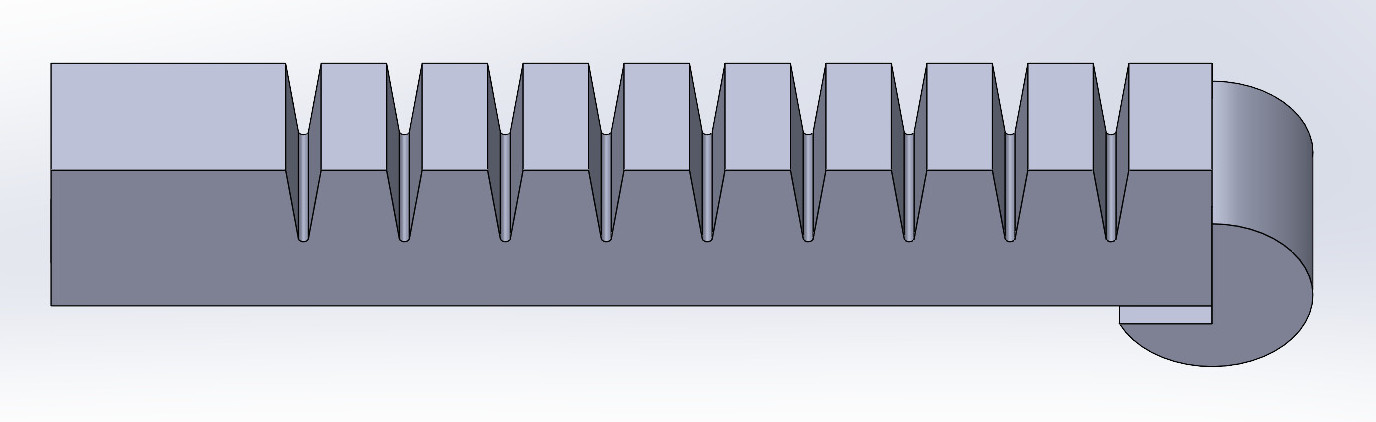}
            \caption{
                }
            \label{fig:fingerCad:OG:full}
        \end{subfigure}
        \begin{subfigure}[b]{0.49\textwidth}
            \centering
            \includegraphics[width=\textwidth]{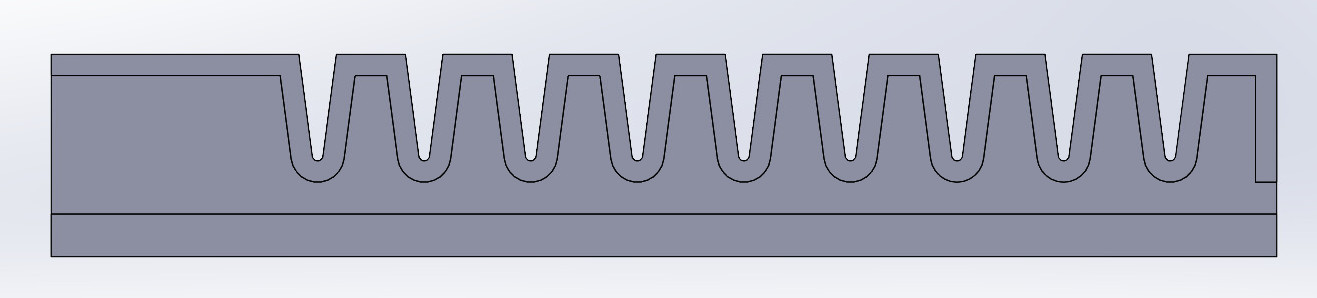}
            \caption{
                }
            \label{fig:fingerCad:OG:cut}
        \end{subfigure}
        \begin{subfigure}[b]{0.35\textwidth}
            \centering
            \includegraphics[width=\textwidth]{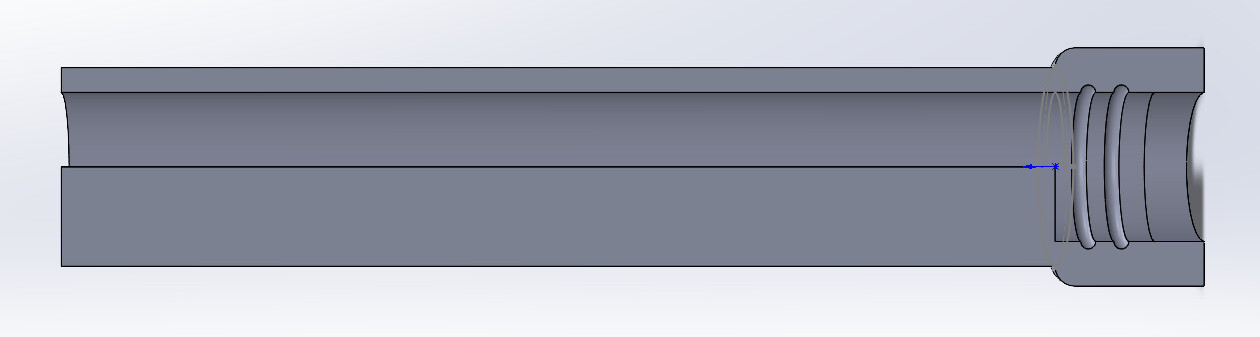}
            \caption{
                }
            \label{fig:fingerCad:oval:cut}
        \end{subfigure}
        \hfill
        \begin{subfigure}[b]{0.12\textwidth}
            \centering
            \includegraphics[width=\textwidth]{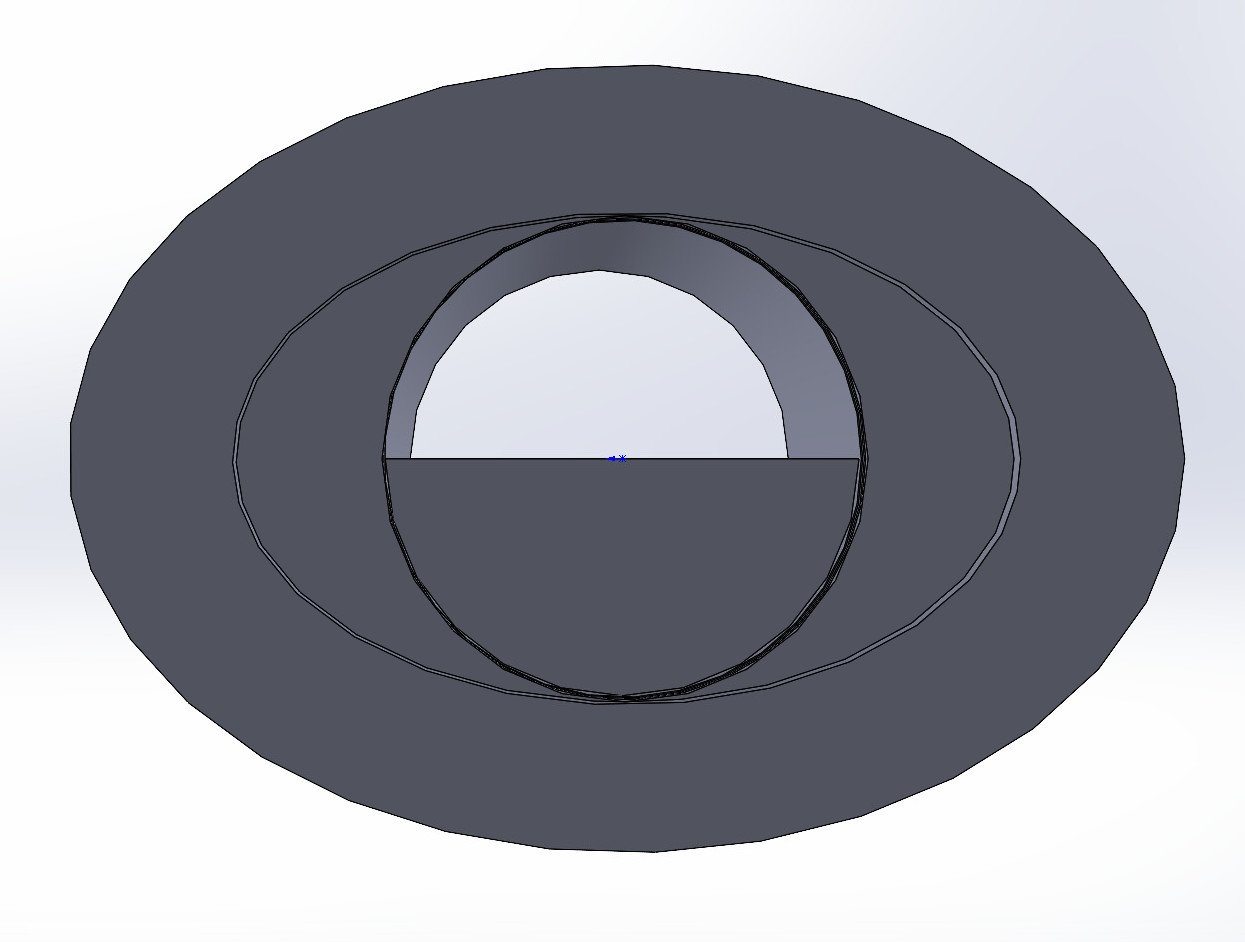}
            \caption{
                }
            \label{fig:fingerCad:oval:base}
        \end{subfigure}
        \caption{The CAD designs of the fingers: a) the first set of moulded fingers with the fingertips attached; b) a cross section of the first set of moulded fingers; c) a cross section of the oval fingers; d) the cavity of the oval fingers.
        }
        \label{fig:fingerCad}
    \end{figure}

    Another pneumatic finger was designed with an oval cross section to encourage bending over its shortest length. A semicircular cavity was placed in the outer side of the finger. It was noticed that due to variations in the curing conditions each finger had a slightly different response to pressure which was rectified by having a separate SMC ITV2050 pressure regulator with its own control curve.
    
    All of the silicone fingers were attached to 3D printed PLA inlets using Sil-Poxy with 4mm pipes glued in to connect them to the pressure regulators. The inlets were also used to hold the fingers on to the chassis.
    
    
    They were later replaced by a set of three 3D printed soft fingers similar manufactured by Inkbit. These had a uniform response to pressure so could be actuated using a single regulator. They were also much lighter and would not deform under their own weight, though they were shorter, which limited the size of graspable objects.
    
\begin{figure}
        \centering
        \begin{subfigure}[b]{0.23\textwidth}
            \centering
            \includegraphics[width=\textwidth]{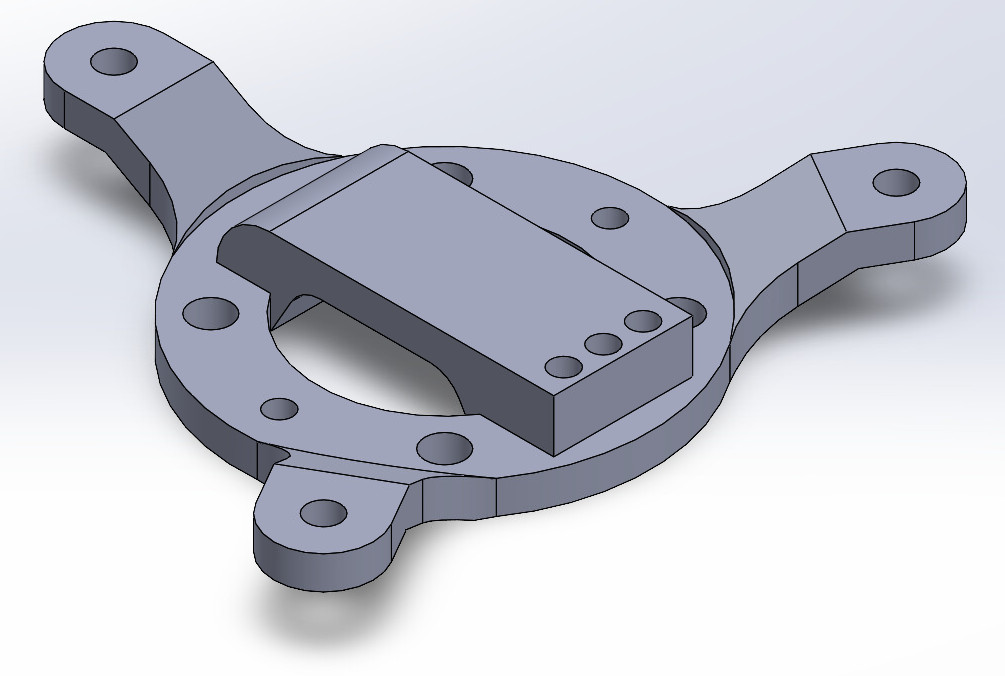}
            \caption{}
            \label{fig:chassis:base}
        \end{subfigure}
        \hfill
        \begin{subfigure}[b]{0.23\textwidth}
            \centering
            \includegraphics[width=\textwidth]{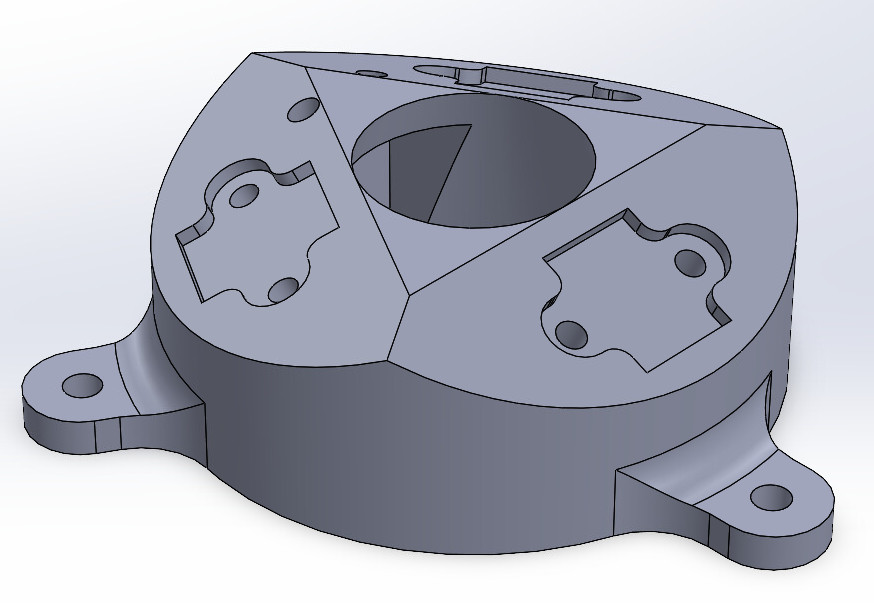}
            \caption{}
            \label{fig:chassis:top}
        \end{subfigure}
        \caption{The chassis was 3D printed from PLA in two parts to allow access to the motor: a) The base plate of the chassis where the motor was mounted; b) The cover of the chassis that supported the fingers on the sloped exterior.}
        \label{fig:chassis}
    \end{figure}

    
    The chassis was 3D printed from PLA in two parts. The base plate had bolt holes to attach the servo motor \customfigBracket{fig:chassis:base}. The cover had mount-points for each finger recessed into slanted edges that were designed for easy replacement \customfigBracket{fig:chassis:top}. The palm attached directly to the servo motor through a hole in the cover.
    
    The control system consisted of an Arduino, a Raspberry Pi and a PC linked together using a ROS framework. The Raspberry Pi and Arduino provided a 0-5V pulse-width-modulated output to each pressure controller, a serial output to the motor, and toggled the 24V supply to the valve for the vacuum pump, while the PC was used as the controller.
    
    When the moulded fingers were used, the Arduino was programmed to map a single linear input from the PC to the three separate voltage ranges for the pressure controllers. The ranges were manually set to align the responses, allowing the fingers to bend simultaneously. When using the 3D printed fingers, a single actuation single going to all three fingers in parallel was sufficient.

\section{Experimental Study}
    
    Grasping and rotating were tested with a variety of different weights and shapes to ascertain how well the gripper would cope with manipulating real-world objects. The sizes of objects used were restricted to those within the grasping capabilities of the fingers. They had to be small enough to be grasped and below 80g for the fingers to be able to lift them.
    
    The tests were carried out by performing the full grasping process on each object five times. All of the objects were tested with both the oval moulded fingers and the 3D printed fingers \textit{(Table \ref{tbl:weights})}. Each one was grasped from above, rotated 180$^\circ$ to above the gripper, dropped on the palm, re-oriented, re-grasped and placed back down. Because the whole sequence had to be completed without error for an object to be deemed as having been successfully manipulated, the individual stages were carefully observed to ascertain how a particular object's properties affected the process. If, for example, the process repeatedly failed at a particular stage, the test would be continued at the start of the subsequent stage.
    
    
    \begin{table}
        \begin{center}
            \begin{tabular}{|| l | r ||}
                Object & Mass (g) \\
                \hline
                Styrofoam Egg & 1 \\
                Cylindrical container & 33 \\
                Glove & 40 \\
                Tape & 50 \\
                Tennis ball & 62
            \end{tabular}
            \caption{A list of the objects used to test the gripper and their associated weights}
            \label{tbl:weights}
        \end{center}
    \end{table}

\begin{figure}
        \centering
        \includegraphics[width=0.45\textwidth]{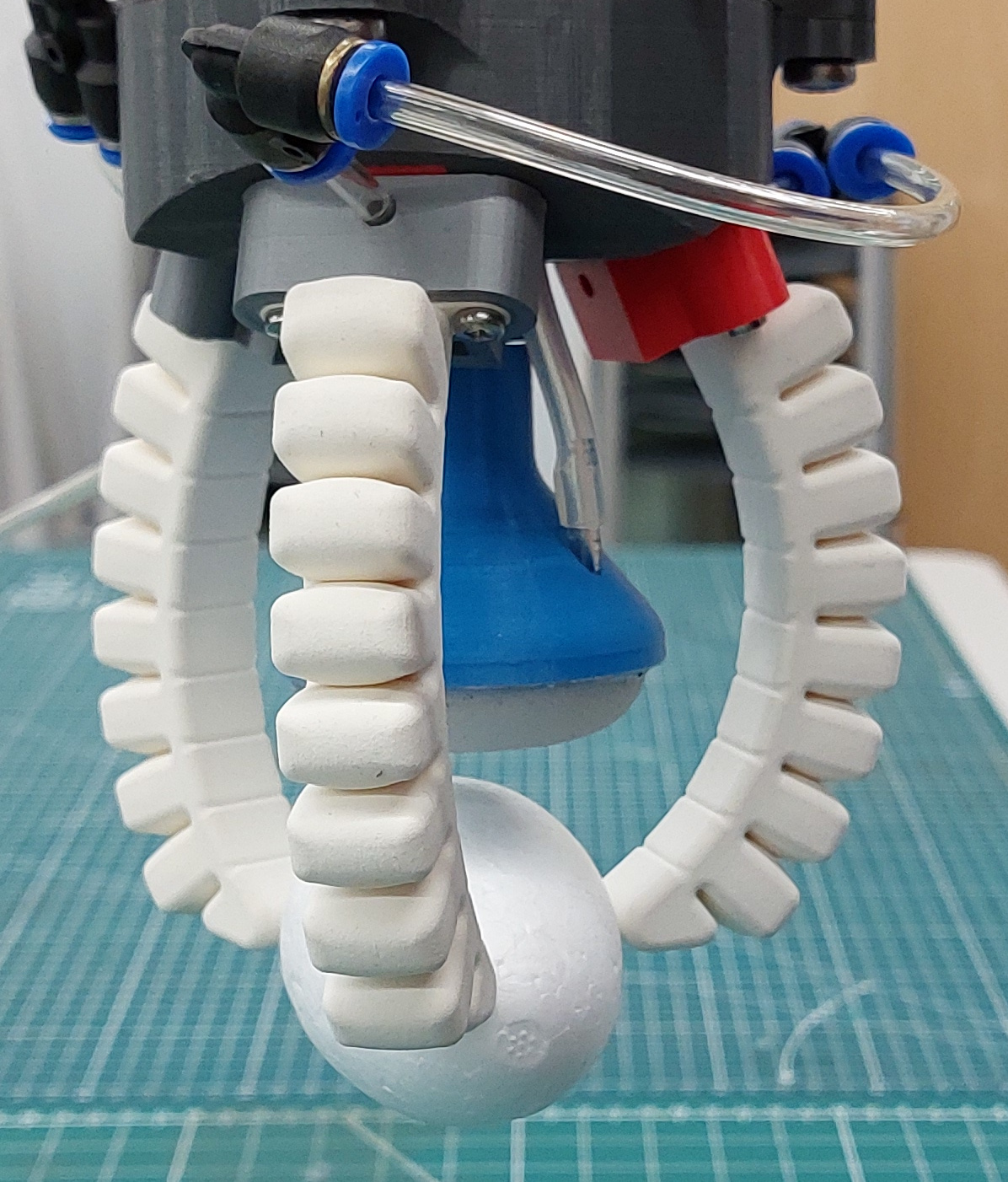}
        \caption{The 3D printed fingers holding the styrofoam egg}
        \label{fig:experiment}
    \end{figure}

    A styrofoam egg weighing 1g was tested first as it was an ideal shape to be grasped and rotated \customfigBracket{fig:experiment}. The moulded fingers had some trouble grasping it as the non-uniform bending would push it out of the centre of the grasping area. They also visibly sagged under their own weight while the gripper was turning face up. The 3D printed fingers grasped and rotated the egg with no issues. Both sets of fingers were able to drop the egg on to the palm easily. The 3D printed fingers held it very close to the palm, so there was little room for deviation. The conical shape the oval fingers that formed when relaxed guided it towards the palm. Once on the palm, the egg could be freely rotated within the servo motor's range of movement. Re-grasping the egg was only possible with the 3D printed fingers as the moulded fingers converged above the egg when flexing.
    
    A cylindrical container was then tested as although it was still an easy shape to grasp, it would have a different weight distribution. Both sets of fingers reliably grasped the container but while rotating it above the palm, the moulded fingers would deform and lose grip. The 3D printed fingers were shorter and could not grasp the cylinder around its center of mass, so the cylinder sometimes twisted out of their grasp. When it was dropped on to the palm it rotated easily. Both sets of fingers were also able to re-grasp the cylinder as it was long enough to pass through the aperture created when the moulded fingers converged. When rotating the palm face down, the moulded fingers were able to support it as they pressed it against the palm. It acted as a fourth point of contact, preventing the cylinder from twisting.
    
    A roll of tape weighing 58g was tested as a heavy object. It was possible to lift with both sets of fingers, but only when held by the gripper with the 3D printed fingers, the tape could be consistently rotated without it dropping. Because it was wide, when it was dropped by the silicone fingers, the differential responses would cause one finger to stay in contact, tipping it as it fell. The roll of tape would then hit the palm at an angle, and either bounce off, falling between the fingers or rest partially on the palm with one edge on a finger.
    
    A rubber glove was also tested to simulate cloth-like items. These were grasped successfully by the 3D printed fingers. The moulded fingers only successfully grasped the glove after pinching it, raising part of the glove off the table to a height at which the cylindrical fingertips could then grasp it. Dropping on to the palm was difficult as the glove would end up draped over the fingers. When it was successfully dropped on to the palm, the glove would fall into the gaps between the fingers prevent it from being turned, even with the suction cup active.
    
    The whole manipulation process was consistently performed successfully on a tennis ball weighing 62g. The fingers interacted well with it as they could reach under it, cage it, and pull it onto the palm before the gripper was rotated to the face up position. This was advantageous as having the center of mass of the object closer to the palm exerted a smaller force on the fingers. As a consequence the ball was never dropped, despite being the heaviest object tested.

\section{Conclusion}
	A soft gripper with rotary palm was designed and implemented. It was tested with a variety of objects and found to be effective at grasping and reorienting them. After successfully grasping, objects weighing up to 62g were easily rotated above 600$^\circ$ per second without slipping. Once on the palm, the objects' dimensions and shape did not effect accuracy - a common occurrence in other fingered grippers. The exceptions were cloth-like objects and large objects which were both obstructed by the surrounding fingers.

	The suction cup on the palm was demonstrably powerful enough to lift objects without using the fingers, which adds extra redundancy to the design. It could also be used in tandem alongside the fingers to increase the maximum lifting capacity.
	
    In future work, the fingers should be made reorientable and retractable to allow for larger objects to be rotated, as the base of the fingers can obstruct larger objects during rotation. The ability to change the angle between the fingers and the centre of the gripper would enable us to address issues such as flexible, cloth-like objects partially falling down in-between the fingers, and wide objects resting on only the palm and not the fingers. A wider spreading of pressurised fingers would also allow wider objects to be grasped. 
    
    It may also be necessary to find a method of preventing the fingers from displacing non-circular objects when re-grasping them. One possible method would be to add contact sensors and individually inflating the fingers until they applied pressure to the object while it is held in place by the vacuum palm. An alternative would be to make the fingers reorient themselves to re-grasp the object in a way that does not disturb it, but that would significantly increase the complexity of the gripper and would require knowledge of the object's shape.
	
	The fingers of the gripper are currently too weak for practical application as they can only reliably lift about 80g. This could be improved by using fibre reinforced silicone or fabric based fingers. Adding electroadhesion would also be a space efficient addition for improving grasping quality.
	

\addtolength{\textheight}{-12cm}   


\section*{Acknowledgment}

Thank you very much to Mish Toszeghi for proof reading.

\printbibliography

\end{document}